\pdfoutput=1

\documentclass[11pt]{article}

\usepackage[]{acl} 

\usepackage{times}
\usepackage{float}

\usepackage{latexsym}
\usepackage{graphicx}
\usepackage{caption}
\usepackage{float}
\usepackage{subcaption}
\usepackage{multirow}
\usepackage{amsmath}
\usepackage{caption} 
\usepackage{array}
\usepackage{booktabs}
\usepackage{tabularx}
\usepackage[table,xcdraw]{xcolor}

\usepackage[T1]{fontenc}

\usepackage[utf8]{inputenc}

\usepackage{microtype}
\usepackage{ifthen}
\newcommand{\DEVELOPMENT}{1} 
\ifthenelse{\DEVELOPMENT = 1}{
	\newcommand{\tz}[1]{\textcolor{purple}{\textbf{TZ:} #1}}	
    \newcommand{\sy}[1]{\textcolor{blue}{\textbf{SY:} #1}}
    }
{
	\newcommand{\tz}[1]{}		
	\newcommand{\sy}[1]{}
}

\title{Show, don't tell -- \\Providing Visual Error Feedback for Handwritten Documents}

\author{
Said Yasin \and Torsten Zesch \\
\small CATALPA -- Center of Advanced Technology for Assisted Learning and Predictive Analytics \\
\texttt{\small said.yasin@studium.fernuni-hagen.de, torsten.zesch@fernuni-hagen.de}
}

\begin{document}
\maketitle
\begin{abstract}
Handwriting remains an essential skill, particularly in education. 
Therefore, providing visual feedback on handwritten documents is an important but understudied area.
We outline the many challenges when going from an image of handwritten input to correctly placed informative error feedback.
We empirically compare modular and end-to-end systems and find that both approaches currently do not achieve acceptable overall quality.
We identify the major challenges and outline an agenda for future research.
\end{abstract}

\section{Motivation}
\begin{figure}[t]
\centering
\includegraphics[scale=0.2]{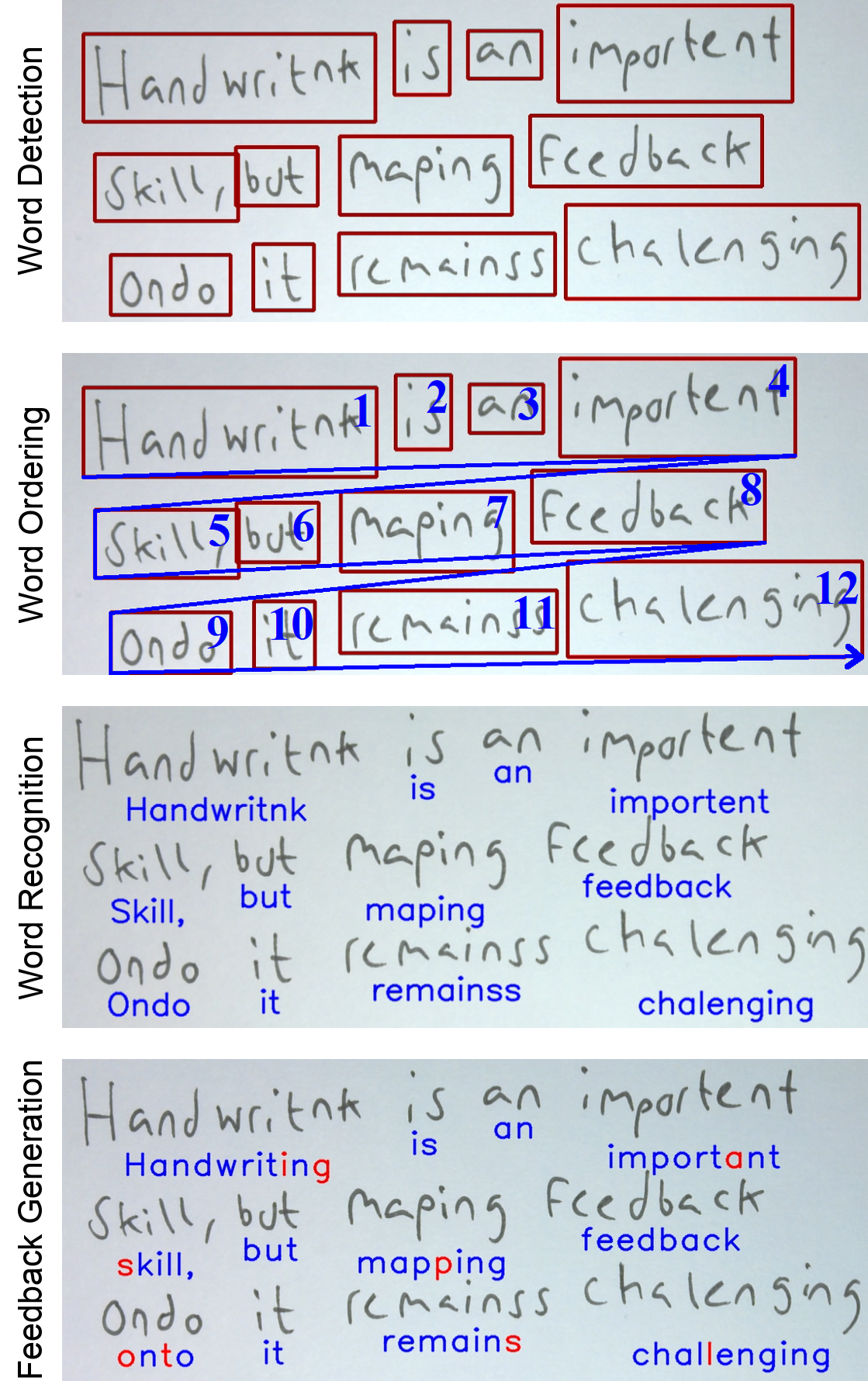}
\caption{Key processing steps in providing visual error feedback for handwriting.}
\label{fig:Feedback_Generation_Process}
\end{figure}

Even in the digital age, handwriting is still the primary writing method used in schools worldwide \cite{Freedman2016}.
It remains an important skill e.g.\ in early language acquisition \cite{ray_relationship_2022}, writing development \cite{feng_roles_2019} or for note taking \cite{MuellerOppenheimer2014}.  
Therefore, the ability to provide digital feedback on handwritten work is essential for bridging the gap between physical documents and current AI technology.
The bottom image in Figure~\ref{fig:Feedback_Generation_Process} shows an example of such feedback.
However, significant challenges remain in word detection, recognition, ordering, and feedback generation, all of which are crucial for providing feedback in handwriting.



\begin{figure*}[t]
\centering
\includegraphics[scale=.16]{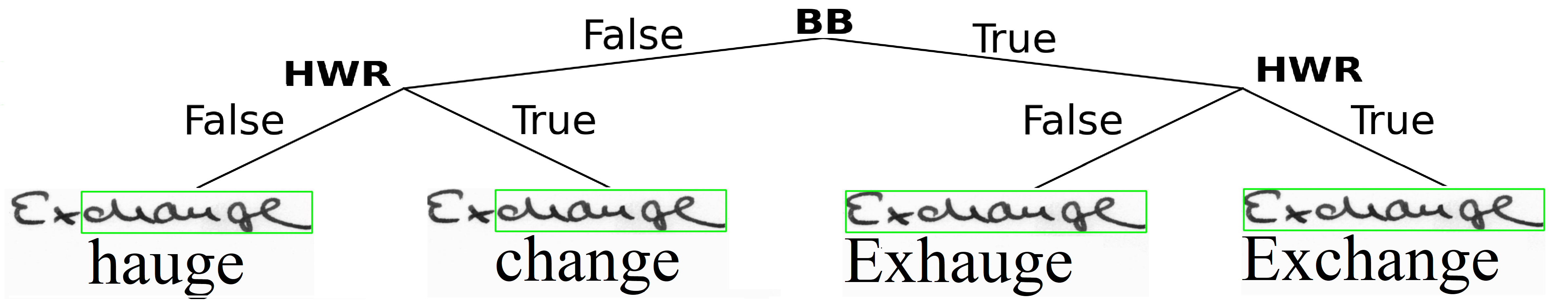}
\caption{Four scenarios in modular word detection and recognition impacting feedback accuracy}
\label{fig:Case}
\end{figure*}

Traditional handwriting recognition (HWR) typically tackle these steps in a modular fashion, where detection, ordering, and recognition are treated as independent, sequential steps \cite{Coquenet2022}. 
While this approach allows for flexible optimization of state-of-the-art text components, it also leads to error accumulation as mistakes propagate between steps \cite{Pengfei}. 
Figure~\ref{fig:Case} illustrates this by showing how erroneous detection influences recognition. 
If a bounding box (BB) is incorrectly positioned, recognition is likely to fail as the correct word cannot be extracted from a truncated BB.
Despite this limitation, modular systems continue to dominate HWR \cite{Coquenet2022}.
Moreover, previous works on feedback systems \cite{KarlssonAkerman2022, Ashini2023} insufficiently analyze the challenges in detection, ordering, and recognition.

To bridge this gap, we provide a systematic analysis of each necessary step and its associated challenges in mapping error feedback onto handwritten text, along with an empirical comparison of modular and end-to-end systems.
These insights are essential for developing applications that enable error annotation on handwriting in educational settings. To validate this and assess practical challenges, we also develop a prototype using a document camera.
We make the source code for the prototype publicly available to foster future research in that area.\footnote{\url{https://github.com/said702/Prototyp_HWR_Feedback}}

\section{Feedback on Handwritten Input}
\label{Feedback_Handwritten}

As we have discussed in the introduction, providing visual feedback on handwritten documents requires a series of steps to be performed (word detection, word ordering, word recognition, and feedback generation) that all come with their own challenges.
In this section, we now discuss each step in more detail.\footnote{Note that, depending on the method used, these steps can occur in a different order. An end-to-end system might directly output correctly ordered text from an image, but still needs to detect where to place the feedback. 
In a modular system, detected words are ordered based on the visual layout before word recognition, or words are recognized in isolation and then ordered using lexical information.}

\subsection{Word Detection}\label{sec:Word_Detection}
Accurate word detection, i.e.\ putting a well-fitting BB around a word in an image, is the essential first step for providing feedback on handwritten documents (see top image in Figure~\ref{fig:Feedback_Generation_Process}).

\begin{figure}[t]
\centering
\includegraphics[scale=0.9]{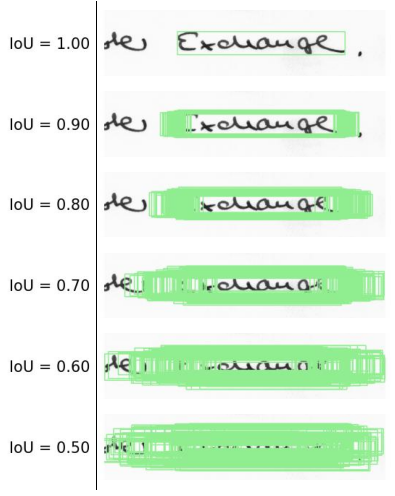}
\caption{Placement of 100 random BBs with specific IoU values.}
\label{fig:BBs_Random}
\end{figure}

Word detection is evaluated using the \textbf{Intersection over Union} (IoU) metric, which quantifies the overlap between two BBs by computing the ratio of their area of intersection to the area of their union:
\begin{equation}
IoU = \frac{\text{Area of Intersection}}{\text{Area of Union}}
\label{eq:iou}
\end{equation}
IoU values may range from 0 to 1, where a higher value indicates greater overlap \cite{Lucas2005}.
Using a threshold $\theta$, which defines the minimum overlap between detected and correct BBs, a system's accuracy can be evaluated.
To illustrate the impact of different $\theta$ values on feedback placement, we plotted 100 random BBs for IoUs between 0.5 and 1 on an image snippet from the IAM dataset (see Figure~\ref{fig:BBs_Random}). 
With decreasing IoU thresholds, BBs deviate markedly from their original positions.
The deviation is more pronounced horizontally due to the rectangular shape of BBs. 
In light of these results, the commonly used IoU threshold of 0.5 for word detection should be reconsidered, as it may lead to inaccurate feedback placement. 
However, maintaining this threshold ensures comparability with other studies \cite{Hakim2024,Ebrahim2024,KarlssonAkerman2022}, but it may lead to an overestimation of accuracy.
For example, \citet{KarlssonAkerman2022} report a detection accuracy of over 95 \% based on an IoU threshold of 0.5. In contrast, \citet{Hakim2024} show that while similarly high accuracy can be achieved at this threshold, performance drops significantly when stricter IoU thresholds up to 0.95 are applied.

\subsection{Word Ordering}
Once words are detected, the next step is to correctly order them to preserve context.
In modular approaches, text ordering is determined independently of detection and extraction, while in end-to-end systems, it is learned in conjunction with these processes.

\begin{figure}[t]
    \centering
    \begin{subfigure}[b]{0.50\textwidth}
        \centering
        \includegraphics[scale=0.2]{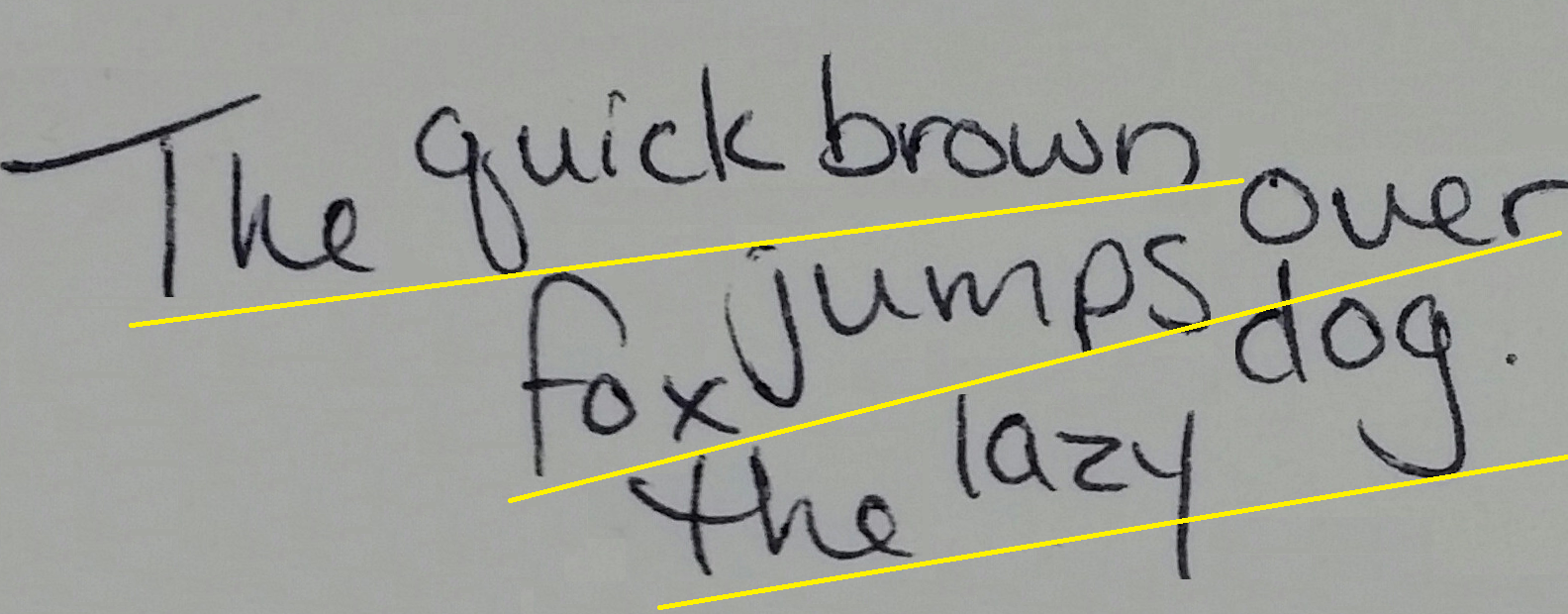}
        \caption{Correct sequencing: ``The quick brown fox jumps over the lazy dog''}
        \label{fig:Correct}
    \end{subfigure}
    \hfill
    \begin{subfigure}[b]{0.50\textwidth}
        \centering
        \includegraphics[scale=0.2]{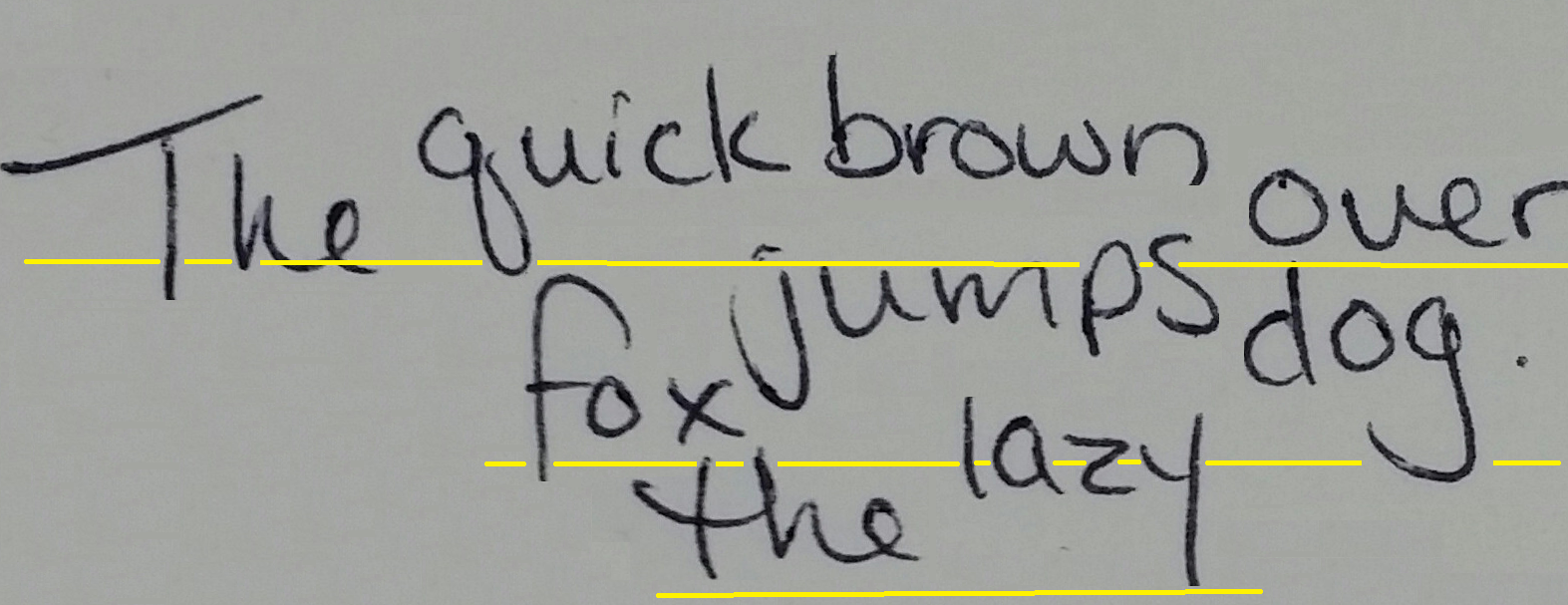}
        \caption{ Horizontal sequencing:  ``The quick brown over fox jumps dog the lazy'' }
        \label{fig:Horizontal}
    \end{subfigure}
\caption{Influence of line recognition on word ordering.} 
    \label{fig:skewed}
\end{figure}

Ordering can either be learned or performed based on heuristics.
In the feedback system by \citet{KarlssonAkerman2022}, word ordering was achieved using a heuristic algorithm that first calculates the average box height, groups boxes based on their proximity along the y-axis, and then sorts them by their x-coordinate. 
In contrast, \citet{Coquenet2022} introduced an end-to-end system that directly learns text ordering using a Vertical Attention Network, demonstrating that it outperforms a traditional modular approach.

Rule-based ordering becomes tedious for complex documents and requires prior knowledge of reading order, such as whether it is column-based or row-based \cite{Coquenet2022}. 
Additionally, left-to-right heuristics struggle with skewed handwritten text, as illustrated in Figures \ref{fig:Correct} and \ref{fig:Horizontal}.

Word ordering can be evaluated using the Normalized Spearman’s Footrule Distance (NSFD), which measures the normalized sum of absolute positional differences \cite{Vidal2023}, or indirectly via the Bilingual Evaluation Understudy (BLEU) score, which compares n-grams between predicted and reference text \cite{Miles2010}.

\subsection{Word Recognition}
In this step, the handwritten word within the BB is sent to a recognition system that returns the recognized text string.
Performance of this step is usually measured by Character Error Rate (CER), which quantifies the number of incorrectly recognized characters \cite{Clemens2021}.


\citet{KarlssonAkerman2022} review 31 studies on HWR systems using the IAM dataset, reporting CER values ranging from 17\% to 2.9\%, with the TrOCR model from \citet{Minghao2021} achieving the lowest error rate. 
In light of these results, current feedback systems for real-time spelling error detection demonstrated competitive CERs, with \citet{KarlssonAkerman2022} achieving 3.4\% using TrOCR on IAM test data and \citet{Ashini2023} reporting 3.4\% on the IAM and SROIE datasets through a Transformer-based extraction approach. 
In comparison, the end-to-end HWR system based on a Vertical Attention Network \citep{Coquenet2022} capable of processing entire documents while simultaneously recognizing both text and layout, achieved a CER of 3.6\% on the READ 2016 dataset.

\begin{figure}[t]
\centering
\includegraphics[scale=0.83]{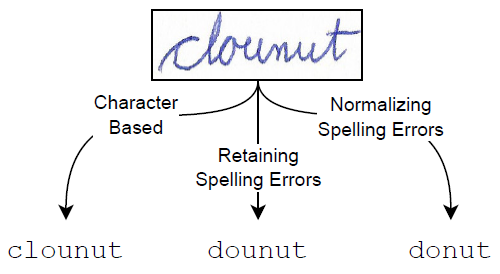}
\caption{HWR output variants from \citet{gold-etal-2023-recognizing}}
\label{fig:Output_Variants}
\end{figure}

While low single digit CER values seem low, this performance level might still severely impact error feedback.
For example, a CER of 5\% means that every 20th character is incorrectly recognized, which might lead to spurious errors being flagged.
However, CER values alone do not tell us whether we are dealing with false positives (spurious errors) or false negatives (missing actual errors).
For example, \citet{gold-etal-2023-recognizing} point out that HWR systems either tend to under-correct or over-correct orthographic errors as shown in Figure~\ref{fig:Output_Variants}.
In the example, a student wanted to write the word \textit{donut}, but spelled it \textit{dounut}.
Due to the shape of the \textit{d} letter, character-based models would recognize \textit{clounut} and add spurious error feedback (as one could argue that the letter looks somewhat like \textit{cl} but is still recognizable as \textit{d} and should not count as a spelling error).
Word-based HWR systems on the other hand over-correct and would recognize \textit{donut} making it impossible to provide error feedback in that case.
The ideal output would be \textit{dounut}, but current HWR systems are not trained to work on this pedagogically useful level.

\subsection{Feedback Generation}
In this study, we focus on error feedback, which involves identifying and marking mistakes in the extracted text, such as spelling and grammatical errors, as well as segmentation errors like split compounds and merged words \cite{Stymne2013}. 
While spelling and segmentation errors are easier to highlight because they affect only individual words or characters, grammar mistakes are more challenging as they often would require showing structural modifications.

\begin{figure}[t]
\centering
\includegraphics[scale=.38]{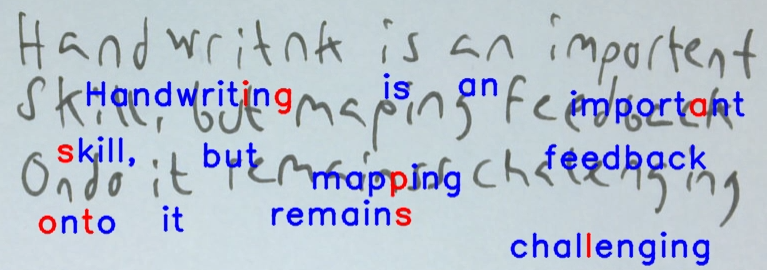}
\caption{Feedback placement with insufficient space between handwritten lines}
\label{fig:tight_handwriting}
\end{figure}

A major challenge for visual feedback is placement, i.e.\ we need to position the feedback in physical space in a way that minimizes distractions.
For example, if there is not enough space between two handwritten lines, the feedback might overlap with the handwriting leading to readability issues, as shown in Figure~\ref{fig:tight_handwriting}.
One possible solution, inspired by language learning support \cite{Rzayev2020}, is to overlay feedback directly on the text. 
However, this hides the original text making it hard for learners to understand the nature of their mistake.
Another approach from language learning support \cite{Mosa2014} is to color-highlight individual letters of the original word, enabling detailed annotation of errors. 
However, this would require locating single characters within a word, while detection approaches rely on word-level BBs.
Also most handwritten datasets are annotated at the word or line level \cite{Marti2002,Kleber2013,Krishnan2021} making it challenging to handle character-level feedback placement.
However, note that (as is shown in the bottom image in Figure~\ref{fig:Feedback_Generation_Process}) we can still provide character-level feedback in form of highlighting the misspelled characters in the word.
What we currently cannot do is directly showing the feedback at the correct character position.

\section{Experimental Setup}
To address the challenges of providing feedback on handwritten documents, as discussed in the previous section, we empirically examine the effectiveness of modular and end-to-end approaches.

\subsection{Data}\label{Data}
We use two datasets: IAM and Imgur5K.

The \textbf{IAM} dataset \citep{Marti2002} is derived from the Lancaster-Oslo-Bergen Corpus and consists of 1,539 scanned handwritten English forms, penned by 657 different writers, with a total of 13,353 isolated text lines and 115,320 words.\footnote{\url{https://fki.tic.heia-fr.ch/databases/iam-handwriting-database}} 
The dataset includes metadata stored in an XML file containing the BB positions and the transcriptions of the handwritten words, which are arranged according to the reading direction, while the entire text is also displayed as printed text in the header of each image.

The \textbf{Imgur5K dataset} \citep{Krishnan2021} comprises 8,177 image pages containing 230,570 words from a diverse range of real-world handwritten samples by nearly 5,300 writers, originally sourced from public posts on Imgur.com.\footnote{\url{https://github.com/facebookresearch/IMGUR5K-Handwriting-Dataset}}
These samples include items such as notes, logos, and various other objects, unlike the form-based samples in the IAM dataset, making Imgur5K particularly challenging.
The dataset also includes metadata in the form of a JSON file, which contains BB positions, transcribed words, and angle information because of variable text orientation.
However, since the words are annotated at the word level and arranged in various layouts, the metadata does not contain explicit ordering information.
In the Appendix (see Figure~\ref{fig:Sample_Datasets}) we provide an example image from each dataset.

\paragraph{Resizing}
All images are resized to 800×800 pixels, as our experiment showed that this leads to improved results (see Appendix~\ref{Resizing} for details).
No further preprocessing is applied, as tests with OpenCV\footnote{\url{https://docs.opencv.org/4.x/d7/dbd/group__imgproc.html}} processing methods showed no significant overall benefit.

\subsection{Approaches}
In this study, we compare the performance of PGNet, Tesseract, EasyOCR, and the Handprint library (using Google Cloud Vision API). 
We limit ourselves to systems capable of generating BBs, a feature lacking in many deep learning systems \cite{Curtis2023} and crucial for feedback placement.

\paragraph{End-to-end system} We selected PGNet \cite{Guangming2021}, an end-to-end OCR system from \mbox{PaddleOCR}, due to the limited availability of research and open-source end-to-end HWR systems \cite{Coquenet2022}, as well as its efficiency in avoiding complex, time-consuming steps like non-maximum suppression and region of interest.

The basic idea of PGNet\footnote{\url{https://github.com/PaddlePaddle/PaddleOCR/blob/main/doc/doc_en/algorithm_e2e_pgnet_en.md}} is to combine text detection and recognition by sharing CNN features and training both components simultaneously. The system learns text ordering by organizing center points via directional offsets, enabling recognition in non-traditional reading directions.

\paragraph{Modular systems}
We selected Tesseract and EasyOCR for their notable performance in comparative studies \cite{Qurban2024}.

Tesseract\footnote{\url{https://github.com/tesseract-ocr/tesseract/}} detects text using a binarization algorithm, followed by Convolutional Neural Networks (CNNs) for character recognition and Recurrent Neural Networks with Long Short-Term Memory for sequence processing \cite{Kongkla2022}. 

EasyOCR\footnote{\url{https://github.com/JaidedAI/EasyOCR}} detects text using Character Region Awareness for Text Detection (CRAFT) \cite{Youngmin2019}, which relies on CNNs, and utilizes a Convolutional Recurrent Neural Network (CRNN) \cite{Baoguang2015} for extraction.

\paragraph{Cloud}
For comparison, we also use the Handprint library through Google Cloud Vision API, whose specific methods remain undisclosed \cite{Kongkla2022}.

\subsection{Evaluation}
\label{Metrics}
\paragraph{Word Detection}
We evaluate word detection using precision, recall, and F1 score, considering a BB correctly placed if IoU $\geq$ 0.5 to ensure comparability with related work. To account for the limitations of this threshold (see Section~\ref{sec:Word_Detection}), we also plot the distribution of all IoU values.

\paragraph{Word Recognition}
We evaluate word recognition using the CER. Only successfully detected words are considered, ensuring that undetected words do not contribute a CER of 1.0,
\paragraph{Word Ordering}
We evaluate word ordering using the BLEU score and NSFD metrics. To exclude recognition errors, we use the system's proposed order and replace predicted words with the ground truth.

\section{Empirical Analysis}
This section presents and analyzes the performance of modular and end-to-end systems to understand how their detection and extraction capabilities vary across different datasets. 
This comparison identifies the strengths and weaknesses of each architecture, providing valuable insights for developing an effective feedback system.

\begin{table}[t]
\centering
\setlength{\tabcolsep}{5pt} 
\small
\begin{tabular}{llcccccc} 
\toprule
\textbf{Type} & \textbf{Model} & \multicolumn{3}{c}{\textbf{IAM}} & \multicolumn{3}{c}{\textbf{Imgur5k}} \\
 &  & \textbf{P} & \textbf{R}  & \textbf{F} & \textbf{P} & \textbf{R}  & \textbf{F} \\
\midrule
E2E & PGNet     & .83 & .60 & .70 & .62 & .46 & .53 \\
Modular    & EasyOCR   & .74 & .58 & .65 & .58 & .41 & .48 \\
Modular    & Tesseract & .78 & .74 & \bf{.76} & .30 & .28 & .29 \\
Cloud    & Handprint & .65 & .65 & .65 & .67 & .67 & \bf{.67} \\
\bottomrule
\end{tabular}
\caption{Word detection results}
\label{table:PrecisionRecall}
\end{table}

\subsection{Word Detection}
\begin{figure}[t]
    \centering
    \begin{subfigure}[b]{0.5\textwidth}
        \centering
        \includegraphics[scale=0.35]{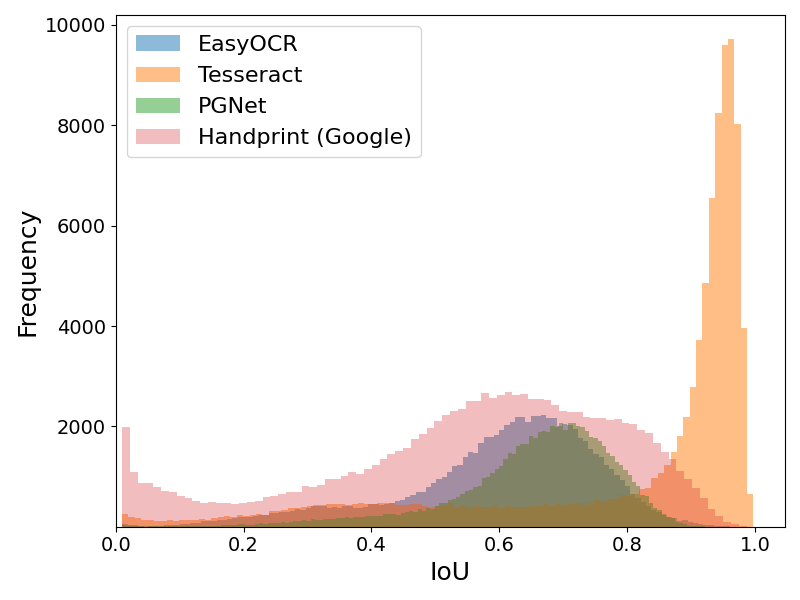}
        \caption{IAM}
        \label{fig:Hist_IAM_IoU}
    \end{subfigure}

    \begin{subfigure}[b]{0.5\textwidth}
        \centering
        \includegraphics[scale=0.35]{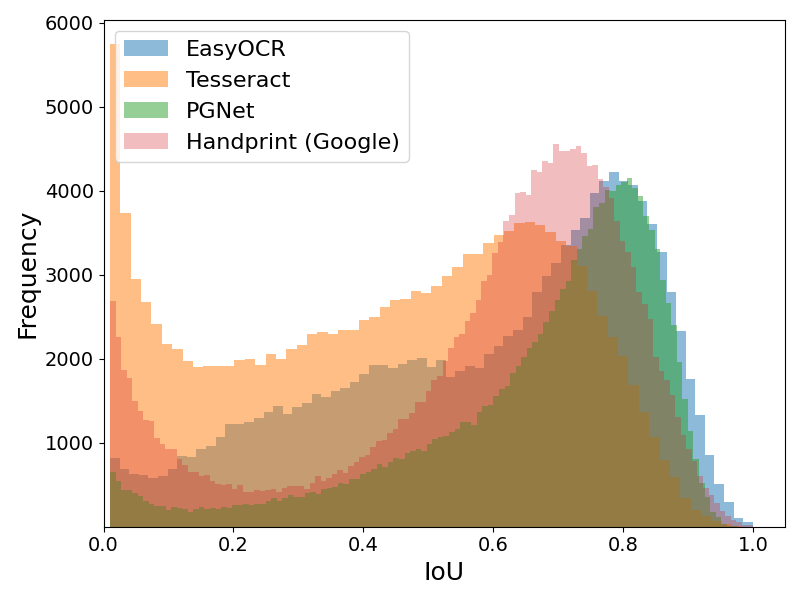}
        \caption{Imgur5k}
        \label{fig:Hist_Imgur5k_Ious}
    \end{subfigure}
    \caption{IoU histograms}
    \label{fig:Hist_Ious}
\end{figure}

Table~\ref{table:PrecisionRecall} shows the results based on an IoU threshold of 0.5, while Figure~\ref{fig:Hist_Ious} illustrates the distribution of all calculated IoUs.
The end-to-end (E2E) system PGNet and Handprint achieve good overall performance, while Tesseract performs well on IAM but poorly on Imgur5k.
However, the IoU histogram for IAM (see Figure~\ref{fig:Hist_IAM_IoU}) reveals unexpectedly high IoUs for Tesseract, possibly indicating data contamination due to undocumented training on the IAM dataset.

\begin{figure}[t]
\centering
\begin{subfigure}[b]{0.25\textwidth}
    \centering
    \includegraphics[scale=1.9]{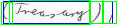}
    \caption{Parentheses and commas}
\end{subfigure}
\begin{subfigure}[b]{0.20\textwidth}
    \centering
    \includegraphics[scale=1.9]{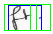}
    \caption{Periods}
\end{subfigure}
\caption{Detection error due to overly detailed BBs illustrated on an IAM snippet (Ground Truth: Blue, Google Cloud Vision API: Green)}
\label{fig:Error_analysis_Google}
\end{figure}

The IoU histogram of Imgur5k (see Figure~\ref{fig:Hist_Imgur5k_Ious}) shows that PGNet and the Handprint library achieve the highest IoUs with similar distributions.
Notably, the Handprint library with Google service exhibits a high frequency of lower IoUs, likely due to its detailed detection of BBs absent from the Ground Truth (see Figure~\ref{fig:Error_analysis_Google}).

Compared to the YOLOv5l detection system trained on the IAM, which achieved 96.1\% recall and 96.2\% mAP with the same IoU threshold of 0.5 \citep{KarlssonAkerman2022}, our systems achieves worse results.
However, their model was trained and tested on IAM (in-distribution), while our evaluation includes out-of-distribution scenarios.

\subsection{Word Ordering}

\begin{table}[t]
\centering
\small
\begin{tabular}{llcc} 
\toprule
\textbf{Type} & \textbf{Model} & \multicolumn{2}{c}{\textbf{IAM}} \\
 &  & NSFD $\downarrow$ & BLEU $\uparrow$ \\
\midrule
E2E & PGNet            & .21 & .18 \\
Modular    & EasyOCR   & .23 & .24 \\
Modular    & Tesseract & \bf{.17} & \bf{.50} \\
Cloud    & Handprint & .21 & .32 \\
\bottomrule
\end{tabular}
\caption{Word ordering results}
\label{table:Ordering}
\end{table}

Since the IAM dataset provides a predefined reading order, we evaluated the models' ordering performance using the NSFD and BLEU metrics (see Table~\ref{table:Ordering}). 
The modular system Tesseract achieves the best ordering score, likely influenced by its higher recall (see Table~\ref{table:PrecisionRecall}), which allowed more words to be considered in the evaluation. 
However, this may also be partly due to data contamination as described above.

We provide some example system outputs in Table~\ref{Predicted_Text} in the Appendix.
The examples demonstrate that the word ordering accuracy of all systems is insufficient for generating meaningful feedback.

\subsection{Word Recognition}
\begin{table}[t]
\centering
\small
\begin{tabular}{llrr}
\toprule
& & \multicolumn{2}{c}{CER values $\downarrow$} \\
\textbf{Type} & \textbf{Model} & \textbf{IAM} & \textbf{Imgur5k} \\
\midrule
E2E        & PGNet     & 37      & 50 \\
Modular    & EasyOCR   & 63      & 54 \\
Modular    & Tesseract & 43      & 54 \\
Cloud   & Handprint & \bf{3} & \bf{8} \\
\bottomrule
\end{tabular}
\caption{Word recognition results}
\label{table:CER}
\end{table}

Table~\ref{table:CER} gives an overview of the recognition results.
Our evaluation of extraction performance revealed that the Handprint library with Google service achieved the lowest CERs by a significant margin, which is unsurprising given that the other systems are OCR-specific models.
When focusing solely on the OCR systems, E2E PGNet achieved the lowest CERs, although its IoU distribution on IAM (see Figure~\ref{fig:Hist_IAM_IoU}) closely resembles EasyOCR’s, suggesting improved error propagation handling.

\begin{figure*}[t]
\centering
 \begin{subfigure}[b]{0.30\textwidth}
    \centering
    \includegraphics[width=\textwidth]{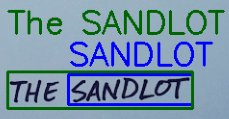}
    \caption{Detection error due to large BBs (EasyOCR)}
    \label{fig:Error_analysis_IoU}
\end{subfigure} \hfill
 \begin{subfigure}[b]{0.32\textwidth}
    \centering
    \includegraphics[width=\textwidth]{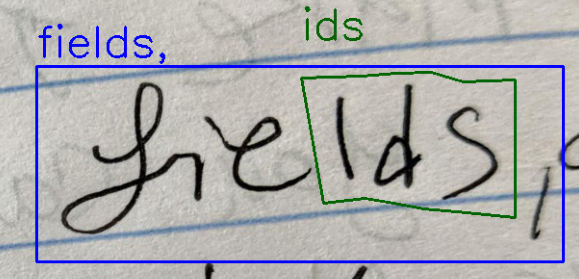}
    \caption{Detection error due to large letter spacing (PGNet) }
    \label{fig:Error_analysis_Detection}
\end{subfigure} \hfill
 \begin{subfigure}[b]{0.32\textwidth}
    \centering
    \includegraphics[width=\textwidth]{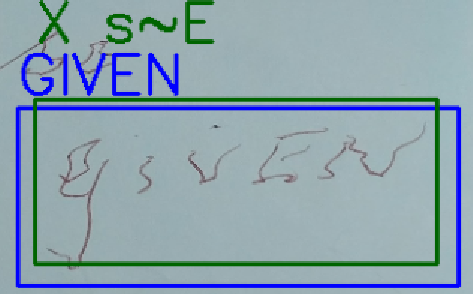}
    \caption{Recognition error caused by unclear handwriting (EasyOCR)}
    \label{fig:Error_analysis_false_HW}
\end{subfigure} \hfill
\caption{Error analysis using Imgur5k snippets (Ground Truth: Blue, Predictions: Green)
}
\label{fig:Error_analysis_combined}
\end{figure*}

However, specific errors contributed to higher CER values across systems. In modular OCR systems, two words were occasionally merged into a single BB, overlapping with the ground truth BB for one word due to the 0.5 IoU threshold (see Figure~\ref{fig:Error_analysis_IoU}).
This mismatch raised the CER, as only one word per BB was expected. 
Similarly, in the E2E system PGNet, large letter spacing caused a comparable issue, where single words were split into two predictions (see Figure~\ref{fig:Error_analysis_Detection}).

Compared to advanced HWR systems, which achieve a word-level CER of 4.9\% on IAM and 9.3\% on Imgur5k \cite{Dmitrijs2022}, the Handprint library with Google service achieves slightly better performance.

\subsection{Summary of Empirical Findings}
Our experiments indicate that neither modular nor end-to-end systems provide sufficient quality for visual feedback on handwriting in classroom settings. While some systems excel in specific categories, their overall performance remains inadequate. Achieving high accuracy across detection, ordering, and extraction requires further research on HWR systems, independent of their architectural approach.

\section{Prototype}
Even if we have concluded from our empirical evaluation that current system performance is probably not ready for classroom use, we still wanted to create a proof of concept where we can experience the feedback quality firsthand.
It is well known that HWR results vary considerably with handwriting style \cite{GoldEtal2021personalization}, so for clearly written block letters performance might already be sufficient.

\begin{figure}[t]
    \centering
    \begin{subfigure}[b]{0.50\textwidth}
        \centering
 \includegraphics[scale=.1]{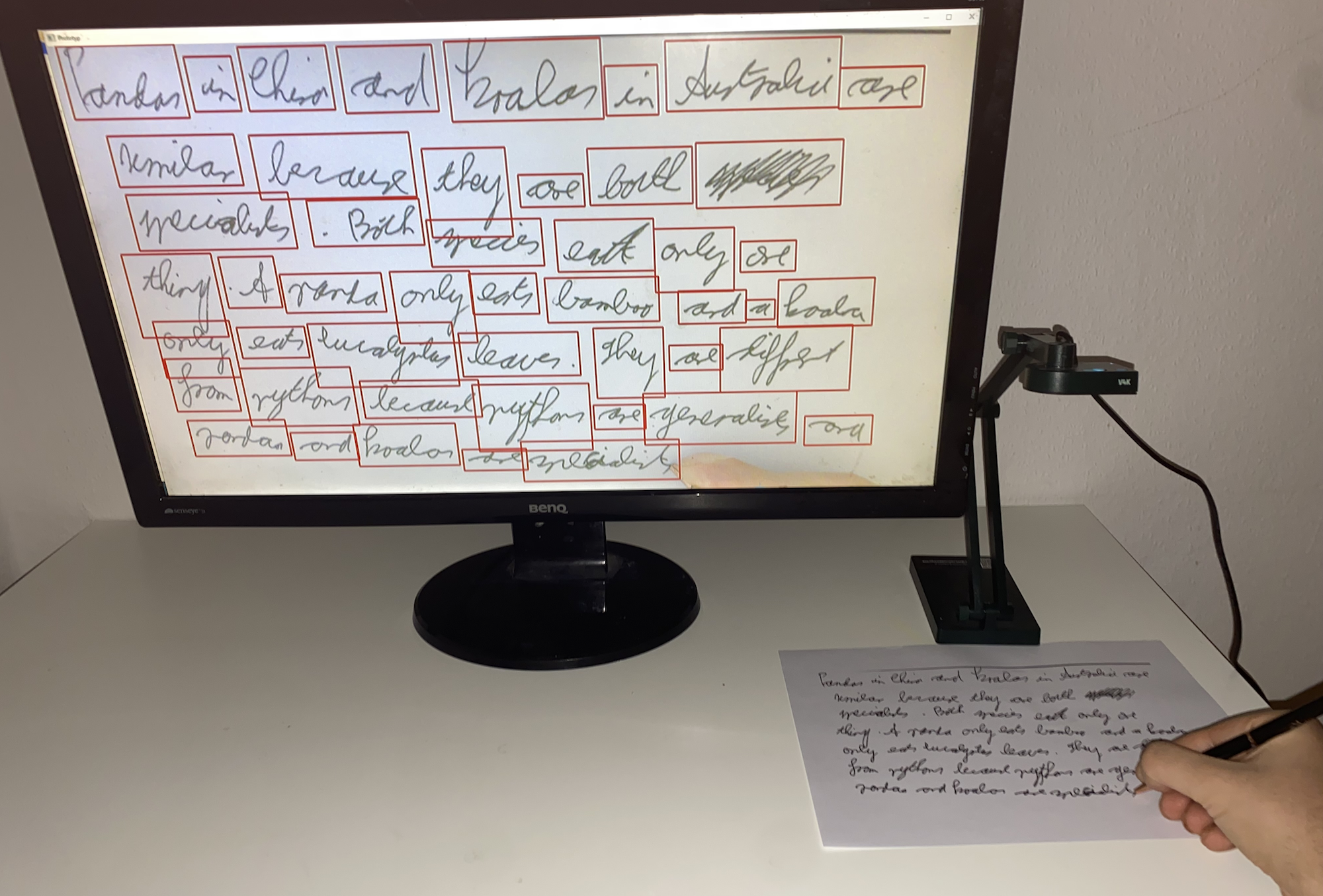}
\caption{Word Detection and Recognition
}
\label{fig:Setup}
    \end{subfigure}
    \hfill
    \begin{subfigure}[b]{0.50\textwidth}
        \centering
     \includegraphics[scale=.43]{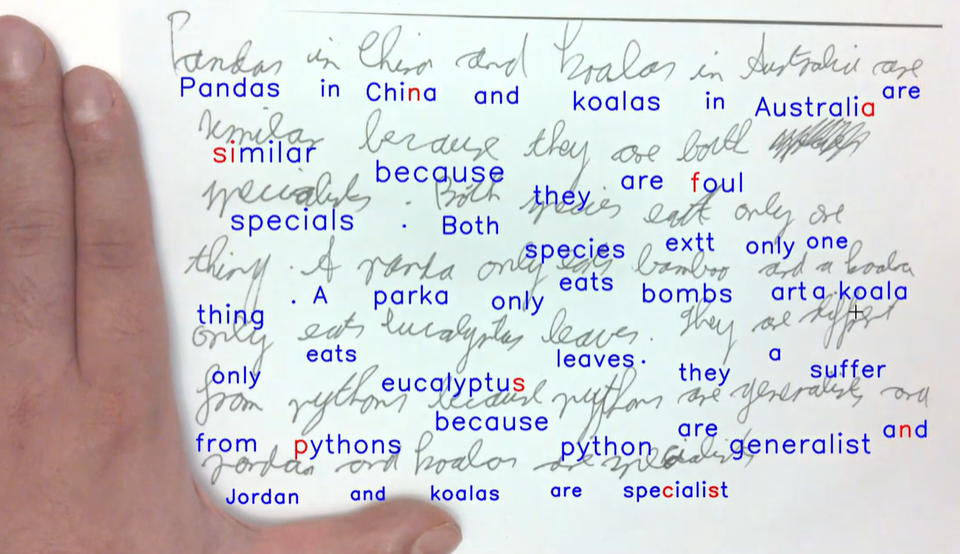}
\caption{Feedback Generation}
        \label{fig:Feedback_Example_prototyp}
    \end{subfigure}
\caption{Prototype setup}
\label{fig:Prototyp}
\end{figure}

We implemented a real-time Python application that uses a document camera to capture handwritten text and display error feedback on the screen.
Initial tests confirm reliable BB detection on document camera images, even adapting to sheet movements within milliseconds (see Figure~\ref{fig:Setup}). 
HWR and subsequent error identification using a spellchecker are also fast with GPU acceleration, allowing near real-time feedback. 
The feedback mapping also works well when there is sufficient space between lines (see Figure~\ref{fig:Feedback_Example_prototyp}). 
When space is insufficient, character-level BBs could help through direct error annotations, particularly for spelling and segmentation errors.
We also observed challenges when the correction differed in length from the original word, making it not always clear which letters to highlight automatically. 

A prototype like this might already be useful for in-class usage, where students could have their writing samples automatically checked at a document camera station within the classroom.
In later iterations of the prototype, we want to replace the screen with augmented reality devices.
Of course, as incorrect feedback might impede learning, the many challenges in improving the overall performance of word detection, ordering and recognition highlighted in this paper, first need to be addressed.

\section{Conclusion \& Future Work}
This study examined key factors in visualizing error feedback for handwritten documents and compared end-to-end with modular systems.

Our analysis indicates that the current state is inadequate for providing digital feedback on handwriting in classroom settings, due to limited system accuracy and many challenges in word detection, ordering, recognition, and feedback generation.
Key issues include the overcorrection by HWR systems and the limited space between handwritten lines.

Despite this limitation, we developed a proof of concept to assess the feedback quality firsthand.
We observed that the prototype was capable of mapping feedback in near real time and, under simplified conditions such as sufficient spacing between lines, placing useful and clearly readable feedback.
Nevertheless, further research is needed to ensure that the prototype can handle the variability in handwriting styles in classroom settings.

\paragraph{Future Work}
Future work should explore methods for determining character-level BBs in handwriting, especially in cursive script, to enable direct error annotation on the original text.
Additionally, further research is needed to mitigate overcorrection in HWR systems to ensure that errors are preserved while achieving a low CER for reliable feedback generation.

Beyond these theoretical considerations, future work should also address practical implementation. 
While our prototype used a document camera with feedback displayed on a screen, the next step is to advance towards augmented reality solutions that utilize a head-mounted display for direct feedback mapping onto handwritten text.

\section*{Limitations}
A key limitation of our approach is that feedback errors are mapped at the word level, which poses challenges when line spacing is tight. 
Character-level BBs could mitigate this issue by enabling annotations directly on the text. 
However, to the best of our knowledge, no large-scale public handwriting dataset provides character-level annotations for word images, which are essential for training a detection model capable of handling cursive handwriting.

Another limitation is that both CER and the ordering metrics were computed only for detected BBs, likely underestimating the recognition error rate by ignoring undetected words and overestimating ordering accuracy, especially when detection outperforms comparable systems.

Experiments were only carried out with English datasets.
Results for other languages are likely worse due to the lower availability of well-performing pre-trained models.

\section*{Ethical Considerations}
Due to variations in handwriting styles and legibility, feedback systems may perform well with some styles but struggle with others, potentially leading to biases and unfair outcomes for individuals whose handwriting is less accurately extracted. 
To ensure fairness, these systems must accommodate a wide range of styles and serve as supportive tools for learning, not penalizing users for handwriting differences.
Another key consideration is data privacy. 
Handwritten data should avoid personal identifiers to prevent exposure of sensitive information. 
Beyond direct identifiers, handwriting style itself can also be considered personal data, as it can reveal an individual's identity if replicated \cite{bhunia2021handwritingtransformers}.

Like all automated feedback systems, there is the risk of providing inaccurate feedback due to system errors which could lead to confusion and worse learning outcomes.
Thus, special care has to be taken to only use such system when a sufficient level of feedback quality can be ensured. 

\bibliography{custom}
\bibliographystyle{acl_natbib}
\clearpage

\appendix

\onecolumn

\section{Examples from IAM and Imgur5k}
\begin{figure}[h]
    \centering
    \begin{subfigure}[b]{0.5\textwidth} %
        \centering
        \includegraphics[scale=0.38]{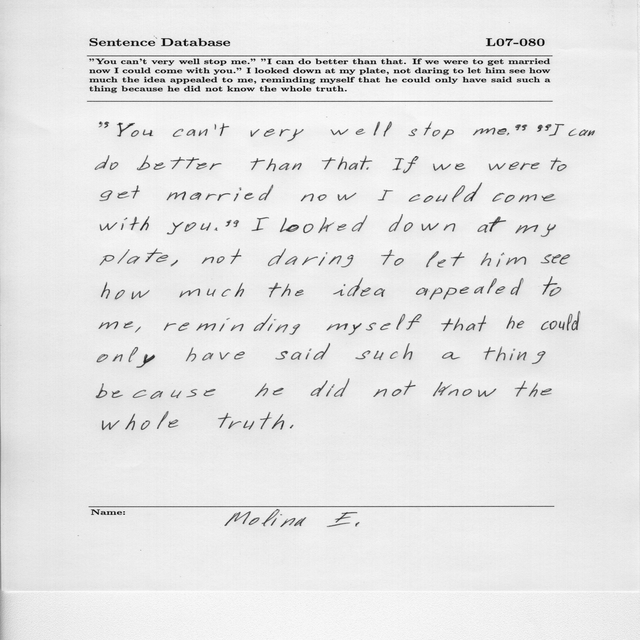}
        \caption{IAM}
        \label{fig:IAM_Example}
    \end{subfigure}
     \hfill 
     \begin{subfigure}[b]{0.5\textwidth} %
        \centering
        \includegraphics[scale=0.22]{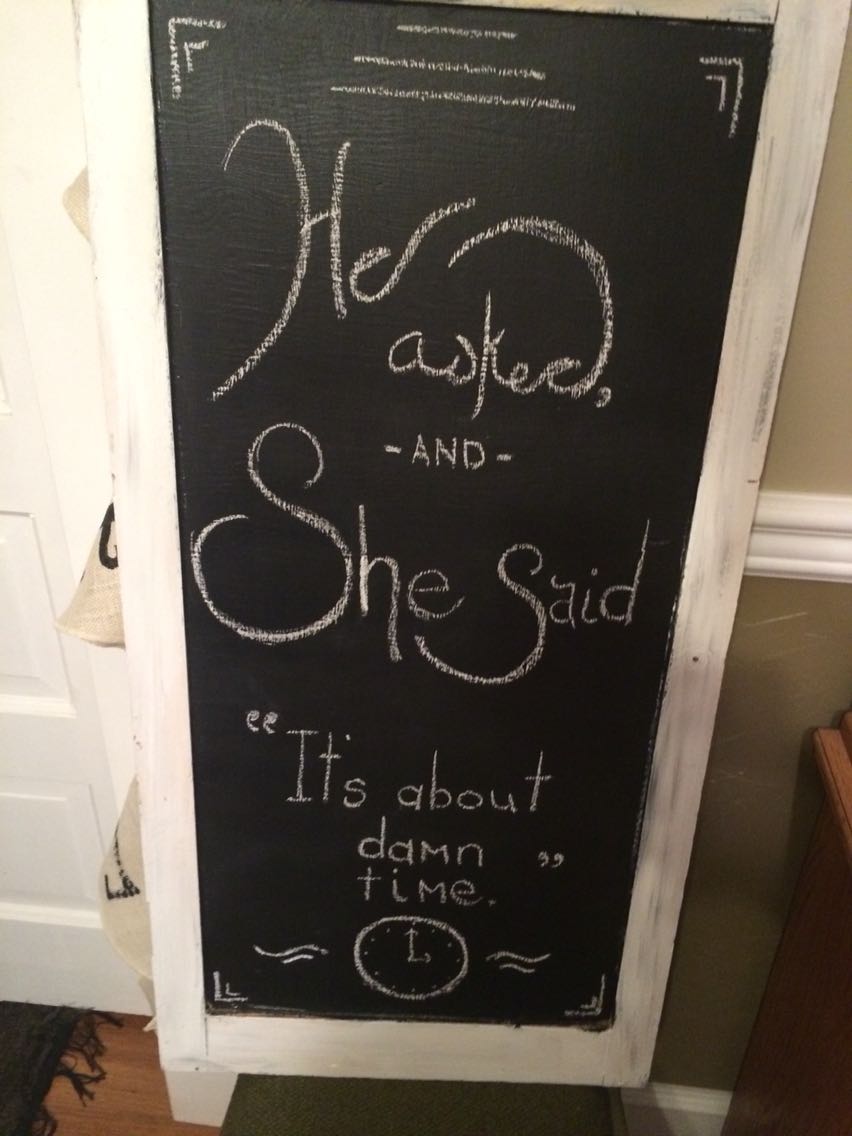}
        \caption{Imgur5K}
        \label{fig:Imgur5K_Example}
    \end{subfigure}
    \caption{Sample images from the utilized datasets}
    \label{fig:Sample_Datasets}
\end{figure}

\clearpage

\onecolumn

\section{Resizing}\label{Resizing}

\begin{table}[h]
\centering
\small
\begin{tabular}{llcccc} 
\toprule
\textbf{Type} & \textbf{Model} & \multicolumn{2}{c}{\textbf{IAM}} & \multicolumn{2}{c}{\textbf{Imgur5k}}  \\
 & & \textbf{w/o resizing} & \textbf{w resizing} & \textbf{w/o resizing} & \textbf{w resizing} \\
\midrule
E2E & PGNet     & .68 & .70 & .52 & .53 \\
Modular    & EasyOCR   & .60 & .65 & .47 & .48 \\
Modular    & Tesseract & .69 & .76 & .29 & .29 \\
Cloud   & Handprint & .63 & .65 & .66 & .67  \\
\bottomrule
\end{tabular}
\caption{Influence of image resizing on word detection results (in terms of F1)}
\label{table:F1}
\end{table}

\begin{figure*}[h]
\centering
\includegraphics[scale=0.5]{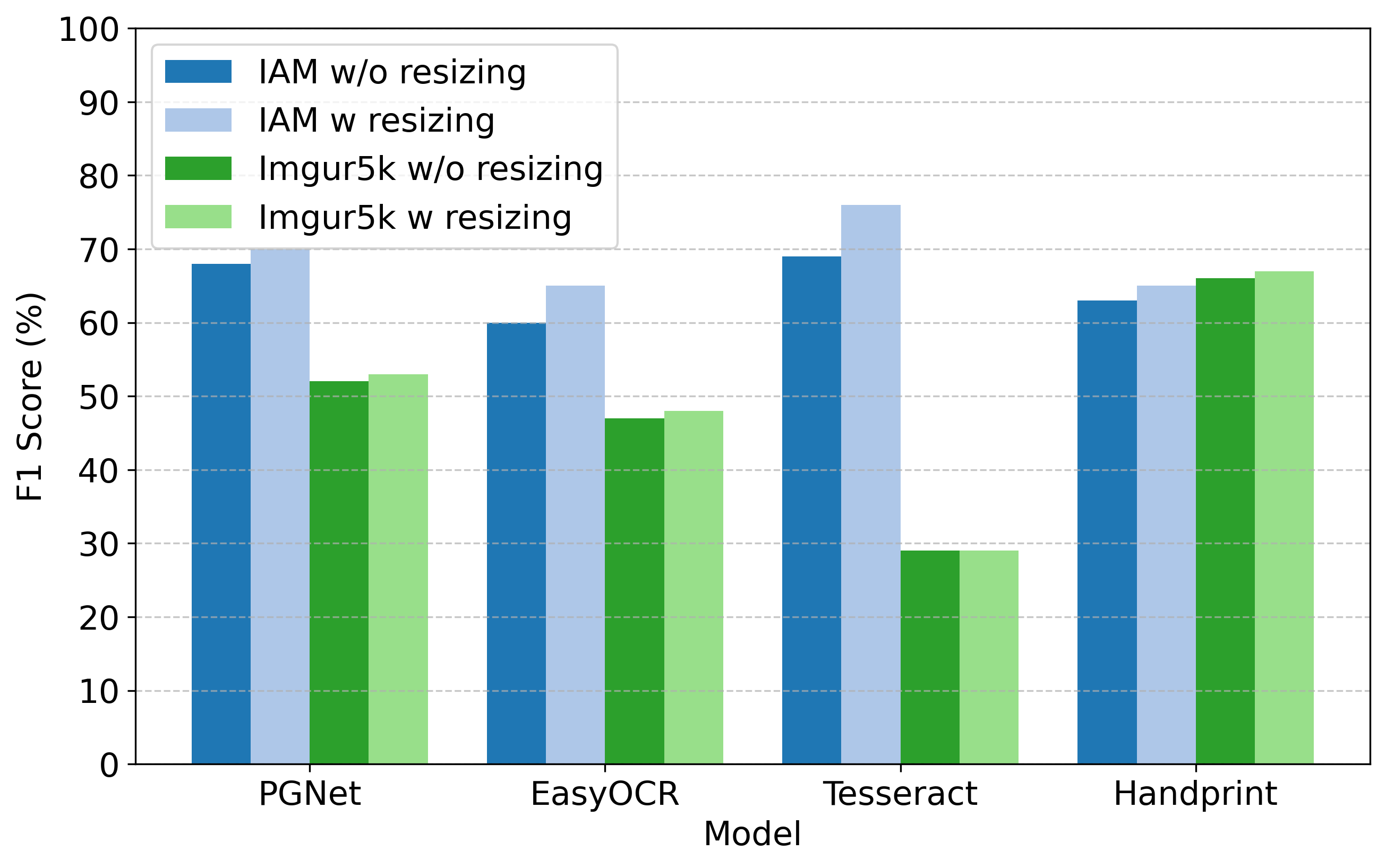}
\caption{Influence of image resizing on word detection results (in terms of F1)}
\label{fig:Resizing}
\end{figure*}

\clearpage

\section{Analysis: Word detection and ordering}
\begin{table*}[h]
\centering
\begin{tabularx}{\textwidth}{lcccX}
\toprule
 & \textbf{Recall} & \textbf{NSFD} $\downarrow$ & \textbf{BLEU } $\uparrow$ & \textbf{Text}  \\
\midrule
\textit{Original} & - & - & - &
\textit{``You can't very well stop me.'' ``I can do better than that. If we were to get married now I could come with you.'' I looked down at my plate, not daring to let him see how much the idea appealed to me, reminding myself that he could only have said such a thing because he did not know the whole truth.} \\
\midrule
EasyOCR & .69 & .32 & .43 &
You can't very well me do better than that If we married could come with looked down plate not daring to let him see how much the idea appealed to reminding myself that he could only have said such a thing because he did the whole truth stop get not  \\
\midrule
Tesseract & .79 & .26 & .65 &
You can't stop do better than that If we were to get married now I could come with you I looked down at my plate not daring to let him see how much the idea appealed to me reminding myself that he could only have said such a thing because he did not know the whole truth  \\
\midrule
PGNet & .69 & .25 & .26 &
You can't well stop me can very do better than that If were get married could come with looked down at my plate not to daring let him see how much the idea appealed myself that could said only have such thing did he not know the because whole truth \\
\midrule
Handprint (Google) &.76  &.23  & .54 &
You can't very well stop I can do better than that If were to get married I could with you I looked down at plate not daring to let him see how much the idea appealed to reminding myself that he could only have said such thing because he did not know the whole truth\\

\bottomrule
\end{tabularx}
\caption{Example system outputs (detected and ordered words) from an IAM sample (see Figure~\ref{fig:IAM_Example})}
\label{Predicted_Text}
\end{table*}

\end{document}